%% file: main.tex
\documentclass[10pt,twocolumn,letterpaper]{article}

\usepackage[pagenumbers]{cvpr} % To force page numbers, e.g. for an arXiv version

\input{preamble}

\def\paperID{} % *** Enter the Paper ID here
\def\confName{CVPR}
\def\confYear{2026}

\title{Physics-Aware Video Instance Removal Benchmark}

\author{
  \textbf{Zirui Li\textsuperscript{1}}\quad
  \textbf{Xinghao Chen\textsuperscript{2}}\quad
  \textbf{Lingyu Jiang\textsuperscript{1}}\quad
  \textbf{Dengzhe Hou\textsuperscript{1}}\\
  \textbf{Fangzhou Lin\textsuperscript{1,3}}\quad
  \textbf{Kazunori Yamada\textsuperscript{1}}\quad
  \textbf{Xiangbo Gao\textsuperscript{3}}\quad
  \textbf{Zhengzhong Tu\textsuperscript{3,$*$}}\\[6pt]
  \textsuperscript{1}Tohoku University \quad
  \textsuperscript{2}University of Washington\quad
  \textsuperscript{3}Texas A\&M University\\[3pt]
  {\small \texttt{li.zirui.r7@dc.tohoku.ac.jp}}\\[3pt]
  {\small \textsuperscript{*}Corresponding author.}
}

\begin{document}
\maketitle

\input{sec/0_abstract}
\input{sec/1_intro}
\input{sec/2_related_works}
\input{sec/3_dataset}

\input{sec/4_bench}

\input{sec/5_exp}

\input{sec/6_con}

\clearpage
{
    \small
    \bibliographystyle{ieeenat_fullname}
    \bibliography{main}
}

\input{sec/X_suppl}

\end{document}

%% file: preamble.tex
\definecolor{lightblue}{rgb}{0.0, 0.6, 1.0}
\definecolor{darkgreen}{rgb}{0.0, 0.6, 0.0}
\definecolor{lightgreen}{rgb}{0.1, 0.8, 0.1}
\definecolor{lightred}{rgb}{0.7, 0.7, 0.7}
\definecolor{lightgreen}{rgb}{0, 0, 0}
\definecolor{lightlightgray}{rgb}{0.8, 0.8, 0.8}

\usepackage{amsmath}  % For mathematical symbols
\usepackage{upgreek}
\usepackage{array}    % For better array formatting
\usepackage{multirow}
\usepackage{array}
\usepackage{algorithm}
\usepackage{makecell}
\usepackage{graphicx}
\usepackage{algorithm}
\usepackage{algpseudocode}
\usepackage{colortbl}
\usepackage{enumitem}
\usepackage{wrapfig}
\usepackage{mathrsfs}
\usepackage{pifont}
\usepackage{soul}
\usepackage{amssymb}
\usepackage{tcolorbox}
\usepackage{dirtytalk}
\usepackage{xspace}
\usepackage[utf8]{inputenc}

\usepackage{booktabs}

\usepackage{url}
\definecolor{blue}{rgb}{0.21,0.49,0.74}
\definecolor{red}{rgb}{0.8, 0.2, 0.2}
\definecolor{green}{rgb}{0, 0.5, 0}
\definecolor{yellow}{RGB}{218, 160, 109}
\definecolor{best}{HTML}{2E7D32}    % 深绿色 (Best)
\definecolor{second}{HTML}{A5D6A7}  % 浅绿色 (Second Best)

\usepackage[pagebackref,breaklinks,colorlinks,citecolor=blue]{hyperref}
\usepackage[capitalize]{cleveref}
\crefname{section}{Sec.}{Secs.}
\Crefname{section}{Section}{Sections}
\Crefname{table}{Table}{Tables}
\crefname{table}{Tab.}{Tabs.}
\crefname{figure}{Fig.}{Figs.}
\Crefname{figure}{Figure}{Figures}
\crefname{appendix}{App.}{Apps.}
\Crefname{appendix}{Appendix}{Appendices}

%% file: sec/0_abstract.tex
\begin{abstract}
\textbf{Video Instance Removal (VIR)} requires removing target objects while maintaining background integrity and physical consistency, such as specular reflections and illumination interactions. Despite advancements in text-guided editing, current benchmarks primarily assess visual plausibility, often overlooking the \emph{physical causalities}---such as lingering shadows---triggered by object removal. We introduce the \textbf{Physics-Aware Video Instance Removal (PVIR)} benchmark, featuring 95 high-quality videos annotated with instance-accurate masks and removal prompts. PVIR is partitioned into \emph{Simple} and \emph{Hard} subsets, the latter explicitly targeting complex physical interactions. We evaluate four representative methods---PISCO-Removal, UniVideo, DiffuEraser, and CoCoCo---using a decoupled human evaluation protocol across three dimensions to isolate semantic, visual, and spatial failures: \emph{instruction following}, \emph{rendering quality}, and \emph{edit exclusivity}. Our results show that \textbf{PISCO-Removal} and \textbf{UniVideo} achieve state-of-the-art performance, while \textbf{DiffuEraser} frequently introduces blurring artifacts and \textbf{CoCoCo} struggles significantly with instruction following. The persistent performance drop on the \emph{Hard} subset highlights the ongoing challenge of recovering complex physical side effects.
\end{abstract}

%% file: sec/1_intro.tex
\section{Introduction}
\label{sec:intro}

Removing a target instance from a video is a foundational editing capability with direct utility in post-production, privacy protection, robotics simulation, and synthetic data curation.
Compared with image object removal, video instance removal requires consistency over time and consistency with scene physics.
When an object is removed, not only should the object disappear, but its side effects should be updated as well~\cite{motamed2026void, miao2025rose}: reflections in windows, mirror appearances, indirect occlusion patterns, and local illumination cues should all remain plausible.
These requirements expose a major gap between qualitative demos and reliable, comparable evaluation.

Recent video editing and inpainting systems have improved temporal coherence and controllability~\cite{zi2025cococo,wei2025univideo,li2025diffueraser}.
At the same time, methods designed for side-effect-aware removal demonstrate that physics-aware editing is now an explicit research target~\cite{miao2025rose}.
Despite these algorithmic advances, the field lacks a common yardstick: existing evaluations are typically performed on private subsets, inconsistent prompts, and varying resolution constraints, leading to a "closed-world" comparison that obscures true progress.
As a result, it remains difficult to answer basic questions: Which model follows removal instructions best? Which model preserves visual quality over time? Which model minimizes unintended edits outside the target region?

To address this gap, we introduce \textbf{Physics-Aware Video Instance Removal Benchmark}, a task-focused benchmark that standardizes data, protocols, and evaluation.
Our benchmark contains 95 high-quality videos with per-video target segmentation and removal prompts.
We explicitly partition data into \emph{Simple} and \emph{Hard} subsets, where \emph{Hard} clips include stronger interactions with real-world physics (e.g., specular reflections, mirror appearance, and pronounced scene coupling).

A second key contribution is a decoupled human evaluation protocol.
Instead of a single holistic score, we assess three independent dimensions:
(1) \textbf{Instruction Following}, i.e., whether the target is correctly removed;
(2) \textbf{Rendering Quality}, i.e., whether the inpainted result is temporally stable and visually plausible;
(3) \textbf{Edit Exclusivity}, i.e., whether non-target content remains unchanged.
Each dimension uses a 1--4 rubric with explicit criteria, enabling interpretable diagnosis rather than one-number ranking.

We evaluate four representative models under a unified setting: CoCoCo~\cite{zi2025cococo}, UniVideo~\cite{wei2025univideo}, DiffuEraser~\cite{li2025diffueraser}, and PISCO-Removal~\cite{gao2026pisco}.
For PISCO-Removal, we evaluate the variant trained in the PISCO paper~\cite{gao2026pisco} using the ROSE dataset~\cite{miao2025rose}; this version supports more demanding practical configurations, including 720p resolution, portrait orientations, and sequences up to 120 frames.
This cross-model comparison is designed to expose the trade-off between general-purpose inpainting stability and task-specific physical fidelity.

\paragraph{Contributions.}
Our contributions are summarized as follow:
\begin{itemize}[leftmargin=1.05em]
    \item We present a new benchmark for \emph{physics-aware video instance removal}, with 95 high-quality videos, instance-level masks, and removal prompts.
    \item We define a decoupled, interpretable human evaluation protocol with three independent 1--4 metrics and an explicit aggregation rule.
    \item We provide a unified benchmark of four representative methods and establish analysis protocols for overall, per-difficulty, and failure-mode evaluation.
\end{itemize}

\paragraph{Scope and current status.}
This paper focuses on benchmark construction and standardized evaluation rather than proposing a new removal architecture.
Our analysis uncovers a "performance ceiling" where even state-of-the-art models fail to resolve secondary physical interactions, revealing that the primary bottleneck in VIR has shifted from temporal flickering to physical incoherence.

%% file: sec/2_related_works.tex
\section{Related Works}
\label{sec:related_work}

\paragraph{Evolution of Video Inpainting and Instance Removal.} 
Modern video object removal pipelines inherit core spatial advances from image inpainting~\cite{pathak2016context, yu2019freeform, suvorov2022lama} and have evolved through flow-guided propagation~\cite{gao2019fgvc, li2022e2fgvi} and transformer-based temporal modeling~\cite{liu2021fuseformer, zhou2023propainter} to ensure long-horizon consistency. Recent diffusion-based formulations, such as DiffuEraser~\cite{li2025diffueraser}, further enhance realism under complex motion, while ROSE~\cite{miao2025rose} specifically formalizes physical interactions like reflections as first-order constraints. However, while these methods focus on the architectural challenge of texture propagation and physical coupling, our work shifts the focus toward providing a standardized evaluation framework to quantify how well these models actually recover such complex physical causalities.

\paragraph{Text-guided Video Generation and Editing.} 
The field has shifted from general video generation~\cite{rombach2022ldm, ho2022videodiffusion} to highly controllable editing pipelines~\cite{wu2023tuneavideo, geyer2023tokenflow, li2023videop2p}. Recent large-scale models like UniVideo~\cite{wei2025univideo} and CoCoCo~\cite{zi2025cococo} leverage strong generative priors to achieve high-quality restoration and better workflow compatibility. Despite their impressive zero-shot capabilities, these generative models often struggle with "semantic leakage" or fail to strictly respect local masking constraints in instance removal tasks. Unlike these generative frameworks that prioritize visual plausibility, our benchmark emphasizes the decoupling of instruction following and spatial exclusivity to expose these specific failure modes.

\paragraph{Benchmarking and Perceptual Assessment.} 
While general-purpose video benchmarks~\cite{wang2024vbench, huang2024vbenchpp, chen2025ivebench,gao2026pulse} and visual metrics like FID~\cite{heusel2017fid} and FVD~\cite{unterthiner2018fvd} have improved coverage, they often fail to capture high-level physical logic, such as lingering shadows or inconsistent illumination. Existing data infrastructures from segmentation~\cite{perazzi2016davis, xu2018youtubevos} provide annotation principles but do not isolate the entangled dimensions of target compliance and background preservation. Our PVIR benchmark addresses this gap by introducing a physics-aware dataset and a decoupled human evaluation protocol with a 1--4 scoring rubric, providing a more interpretable assessment than conventional automatic metrics.

%% file: sec/3_dataset.tex
\section{Dataset: Physics-Aware Video Instance Removal}
\label{sec:dataset}

\paragraph{Design goal.}
The dataset is designed for one core task: remove a designated instance from a real video while preserving non-target content and maintaining physically plausible side effects.
To enable robust benchmarking, we prioritize high visual quality, diverse scenes, and explicit interaction complexity.

\paragraph{Scale and split.}
The benchmark comprises 95 high-quality sequences, ensuring a balanced distribution between foundational removal tasks and advanced physics-aware challenges.
We organize them into two subsets:
\textbf{Simple} and \textbf{Hard}.
Simple videos (57 videos) usually contain objects with simpler geometry and weaker coupling to scene physics. They serve as a baseline to evaluate a model's fundamental ability to maintain spatial-temporal coherence in the absence of complex physics.
Hard videos (38 videos) include stronger physics interactions, such as mirror reflections, specular highlights, and complex motion-appearance coupling. These cases require the model to not only fill the disoccluded pixels but also maintain physical causality by updating or removing secondary side effects, such as reflections and shadows, that are anchored to the target. Each video includes two annotations:
\begin{itemize}[leftmargin=1.05em]
    \item \textbf{Target segmentation mask}: a high-quality instance mask indicating the object to remove.
    \item \textbf{Removal prompt}: a natural-language instruction that unambiguously refers to the target instance.
\end{itemize}
These two elements define a standardized input interface for all benchmarked models.

\subsection{Collection and Annotation Pipeline}

\paragraph{Data sourcing and pre-filtering.}
Candidate videos are collected from diverse real-world scenes and filtered by minimal requirements on spatial quality, temporal smoothness, and compression artifacts.
To avoid overly synthetic bias, we prioritize clips with natural motion and illumination variation. Our primary sources include the \textbf{Inter4k} dataset~\cite{stergiou2021adapool}, selected for its ultra-high-definition (UHD) clarity and high frame rates, and the \textbf{DAVIS2016} dataset~\cite{perazzi2016benchmark}, chosen for its diversity in object-to-background interactions. Specifically, we curate 45 sequences from Inter4k to ensure high-fidelity rendering assessment, and 50 sequences from DAVIS2016 to leverage its complex motion patterns. All source videos are either licensed under Creative Commons (CC) or are explicitly permitted for non-commercial research use, ensuring a clear and ethical release metadata.

% \paragraph{Instance selection policy.}
% Target instances are strategically selected to encompass a spectrum of geometric and physical interaction complexities. Our taxonomy is organized into three primary categories: 

% \begin{itemize}[leftmargin=1.5em]
%     \item \textbf{Human Subjects (\TODO{XX} instances):} These represent non-rigid, articulated motion. The primary challenge lies in maintaining the integrity of the background during complex limb movements and handling subtle contact shadows.
%     \item \textbf{Animals (\TODO{XX} instances):} This category introduces diverse textures (e.g., fur, scales) and unpredictable motion patterns, testing the model's ability to preserve fine-grained temporal details in disoccluded areas.
%     \item \textbf{Vehicles (\TODO{XX} instances):} Including cars, trains, and boats, these rigid bodies are characterized by high-speed motion and strong environmental coupling. They are the primary source of \emph{specular reflections} on metallic surfaces and complex \emph{wake/track effects} in aquatic or dusty environments.
% \end{itemize}

% By covering these three distinct domains, PVIR ensures a holistic evaluation of Video Instance Removal across varying degrees of structural rigidity and scene interaction.

\paragraph{Mask annotation workflow.}
To ensure pixel-level precision, we employ a multi-stage annotation pipeline. Annotators first utilize the Segment Anything Model 2 (SAM 2)~\cite{kirillov2023segment} to generate initial object tracks across each sequence. By providing sparse point or box prompts on keyframes, SAM 2's memory-based propagation produces coarse masks for the entire video. 
Subsequently, annotators perform manual frame-level refinement to clean up boundaries, particularly in challenging cases involving motion blur or occlusion. To ensure temporal smoothness, we conduct a final quality-control pass focusing on reducing "shape jitter" (flickering boundaries). Any masks exhibiting temporal instability are manually corrected or re-propagated using flow-based consistency checks. This hybrid workflow combines the efficiency of foundation models with the rigor of human verification, yielding the instance-accurate masks required for high-fidelity removal.

\paragraph{Prompt writing workflow.}
Prompts are designed to be specific, concise, and target-disambiguating. To ensure consistency across the benchmark, we adopt a structured template: \texttt{[Action] + [Target Attributes] + [Spatial/Contextual Qualifiers]}. For instance, a PVIR prompt specifies "Remove the silver sedan parked under the flickering streetlamp" instead of a generic command, providing unique semantic grounding that minimizes instruction ambiguity.

About the language policy and disambiguation, all prompts are authored in English using standard descriptive vocabulary. Ambiguous references—such as multiple similar objects in a single frame—are strictly disallowed unless unique qualifiers (e.g., "the person on the far left") are included. For the \emph{Hard} subset, prompts are intentionally augmented with physical context, such as "including its reflection on the water surface," to explicitly signal the expected physics-aware behavior. Each prompt undergoes a cross-verification pass by a second annotator to ensure that the textual description uniquely identifies the instance masked in the ground truth.

\begin{figure*}[t]
    \centering
    \includegraphics[width=\linewidth]{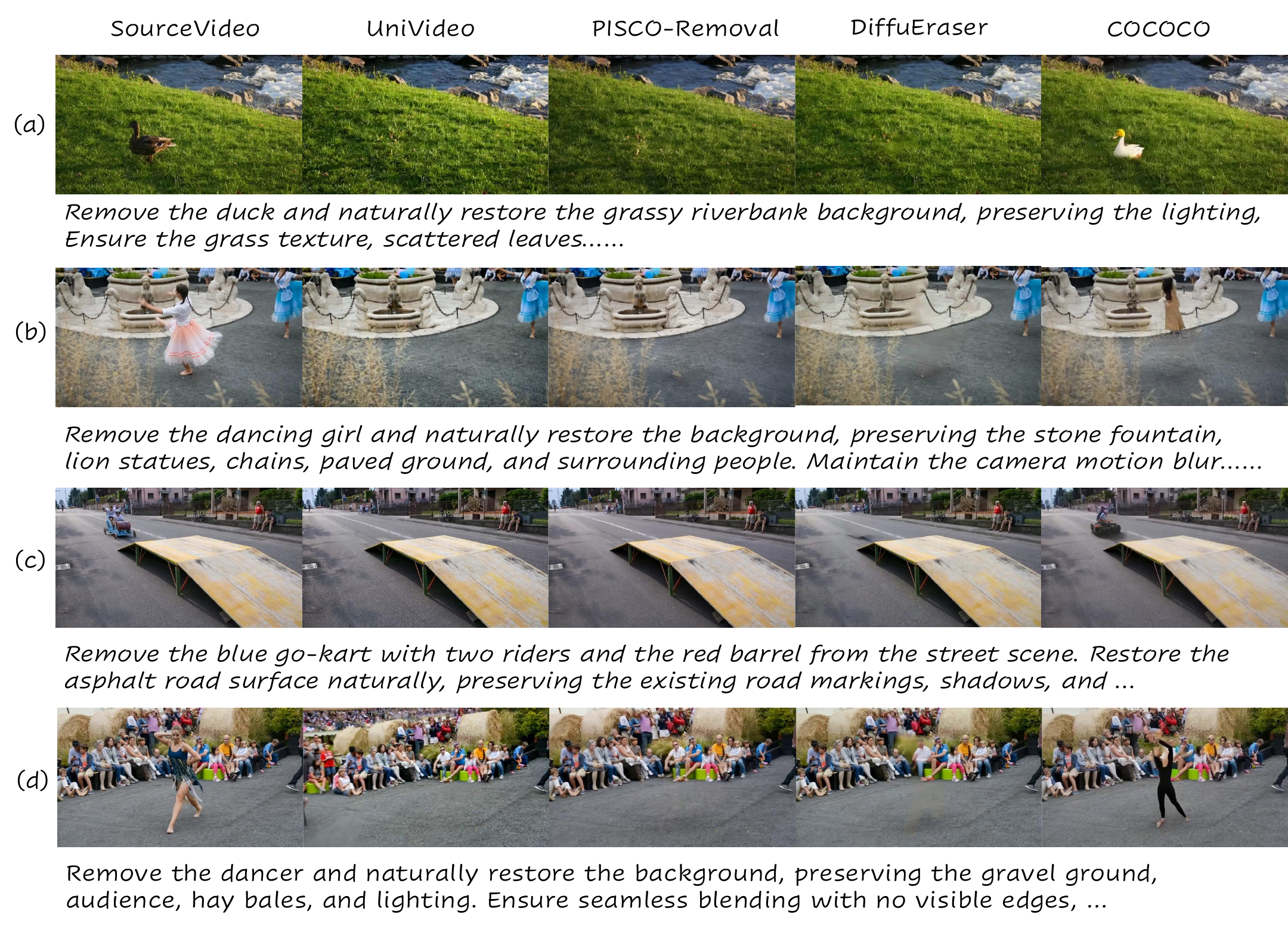}
    \caption{\textbf{Qualitative comparison on the PVIR benchmark.} Each row presents a specific instance removal scenario with its corresponding textual prompt. As a unified model, \textbf{UniVideo} demonstrates strong physics-awareness, successfully removing coupled side effects like ground shadows (e.g., rows b and c); however, it suffers from severe semantic hallucinations, occasionally generating unprompted artifacts to fill the void (e.g., the distorted figure generated in row d). \textbf{PISCO-Removal} consistently achieves clean erasure of both the target and its physical side effects while maintaining high background fidelity. \textbf{DiffuEraser} reliably masks the correct instance but frequently leaves unnatural residual shadows and spatial blurring (rows c and d). Finally, \textbf{CoCoCo} struggles significantly with instruction following, often leaving obvious ``ghosting'' silhouettes or failing to remove the object entirely (e.g., the white duck in row a).}
    \label{fig:dataset_overview}
    \vspace{3mm}
\end{figure*}

\subsection{Difficulty Taxonomy}

\paragraph{Simple subset.}
Simple clips typically contain targets with limited appearance variation and weak side-effect coupling.
Examples include matte surfaces, stable backgrounds, and short-term motion without heavy occlusion.

\paragraph{Hard subset.}
Hard clips emphasize complex light-surface coupling and geometric reconstruction challenges.

\begin{table}[h]
    \centering
    \small
    \caption{\textbf{Summary of the PVIR Benchmark Dataset.} The benchmark comprises 95 high-definition videos, categorized into Simple and Hard subsets based on the complexity of physical interactions (e.g., reflections, shadows, and dynamic fluid wakes).}
    \label{tab:dataset_stats}
    \begin{tabular}{p{0.13\linewidth}p{0.14\linewidth}p{0.14\linewidth}p{0.37\linewidth}}
        \toprule
        Subset & \#Videos & Avg. Frames & Key Properties \\
        \midrule
        Simple & 57 & 81 & weaker interaction and simpler geometry \\
        Hard & 38 & 81 & reflection/mirror/specular and stronger coupling \\
        \midrule
        Total & 95 & 81 & mixed scenes and motions \\
        \bottomrule
    \end{tabular}
\end{table}

% \subsection{Quality Control and Reliability}

% \paragraph{Multi-stage validation.}
% We perform three rounds of validation:
% (1) annotation integrity check,
% (2) prompt-target consistency audit,
% (3) difficulty-label review.
% Disagreements are resolved via re-annotation and adjudication.

% \paragraph{Reliability protocol.}
% To ensure the high fidelity of our benchmark, we implement a multi-stage verification pipeline. A random subset of 20\% of the video sequences is independently reviewed and cross-annotated by three senior researchers for calibration. 

% We measure the inter-annotator agreement using the Mean Intersection-over-Union (mIoU) for masks and \emph{Cohen’s Kappa} coefficient for textual prompts. Our initial assessment yielded an mIoU of 0.92 and a Kappa score of 0.88, indicating strong consensus. 

% During the quality-control pass, approximately 15 of the clips were flagged for relabeling—primarily due to minor temporal boundary drifting in the \emph{Hard} subset or ambiguous descriptors in the prompts. Following two rounds of iterative refinement, the final error rate was suppressed to below 2\% across all dimensions.

%% file: sec/4_bench.tex
\section{Benchmark Protocol}
\label{sec:bench}

\paragraph{Task definition.}
Given an input video $V$, a target mask sequence $M$, and a removal prompt $P$, the model outputs an edited video $\hat{V}$ where the target instance is removed.
The benchmark requires:
(1) complete target removal alongside its associated physical derivatives,
(2) high-quality temporally consistent rendering,
(3) minimal unintended changes outside the edited region.

\paragraph{Evaluated models.}
We evaluate four representative methods:
\textbf{CoCoCo}~\cite{zi2025cococo}, \textbf{UniVideo}~\cite{wei2025univideo}, \textbf{DiffuEraser}~\cite{li2025diffueraser}, and \textbf{PISCO-Removal}~\cite{gao2026pisco}, where PISCO-Removal is a WACE~\cite{jiang2025vace}-like model fine-tuned with the ROSE dataset~\cite{miao2025rose} that support 720p, portrait videos, and up to 120 frames~\cite{gao2026pisco}.

\subsection{Unified Inference Setup}

\paragraph{Input/output interface normalization.}
Each model receives the same semantic input triplet $(V, M, P)$ and outputs a completed video sequence.
We standardize video decoding, prompt formatting, and export codecs to minimize evaluation noise from non-model factors.

\paragraph{Method-specific adaptation policy.}
To eliminate performance bias stemming from disparate input constraints, all evaluated baselines are unified under a standardized configuration: 720p resolution at 81 frames. Unlike previous benchmarks that often resort to downsampling or temporal truncation, our protocol ensures that every model is tested at its maximum practical capacity. We utilize the officially released checkpoints for each method, ensuring that any observed performance gaps are intrinsic to the models' architectures rather than artifacts of suboptimal parameter tuning. This high-resolution, long-duration setup serves as a rigorous stress test for temporal-physical consistency in Video Instance Removal.

\subsection{Human Evaluation Protocol}
\label{subsec:human_eval}

\paragraph{Decoupled scoring principle.}
All three dimensions are scored \emph{independently}; a score in one dimension must not influence another.
Each dimension uses a 1--4 ordinal rubric, where higher is better.

\paragraph{(1) Instruction Following (IF).}
\textbf{Core question:} Does the edited video correctly satisfy the removal instruction?
For this benchmark, judges check whether the specified target instance is removed, and whether visible remnants (edges, fragments, obvious traces) remain.

\paragraph{(2) Rendering Quality (RQ).}
\textbf{Core question:} Is the filled content visually plausible over space and time?
Judges inspect naturalness, sharpness, temporal stability (flicker/jitter), and physical plausibility of motion/appearance.

\paragraph{(3) Edit Exclusivity (EE).}
\textbf{Core question:} Did the model only perform the requested removal?
Judges verify that non-target regions preserve original content, including background structures, lighting appearance, and unrelated objects.

\begin{table*}[t]
    \centering
    \small
    \caption{\textbf{Decoupled Human Evaluation Rubric.} Each of the three dimensions---Instruction Following (IF), Rendering Quality (RQ), and Edit Exclusivity (EE)---is scored independently on a 1--4 scale, enabling granular diagnosis of model failure modes.}
    \label{tab:rubric_full}
    \begin{tabular}{lp{0.28\linewidth}p{0.28\linewidth}p{0.28\linewidth}}
        \toprule
        Score & Instruction Following & Rendering Quality & Edit Exclusivity \\
        \midrule
        Score 1 & \textit{Target not removed or unrelated edit} & \textit{Severe artifacts; unusable output} & \textit{Uncontrolled edits across scene} \\
        \addlinespace[5pt]
        Score 2 & \textit{Partial removal with obvious residual traces} & \textit{Obvious distortion/flicker; poor temporal consistency} & \textit{Multiple non-target regions altered} \\
        \addlinespace[5pt]
        Score 3 & \textit{Target mostly removed with minor artifacts} & \textit{Moderate quality degradation but viewable result} & \textit{Minor non-target changes but structure mostly preserved} \\
        \addlinespace[5pt]
        Score 4 & \textit{Target precisely removed with no obvious residuals} & \textit{High visual quality and stable temporal consistency} & \textit{Non-target regions preserved with only negligible differences} \\
        \bottomrule
    \end{tabular}
    \vspace{3mm}
\end{table*}

\paragraph{Aggregation.}
Let $s_{d}^{(i,r)} \in \{1,2,3,4\}$ denote the score for video $i$, rater $r$, and dimension
$d \in \{\text{IF},\text{RQ},\text{EE}\}$.
We first average over raters:
\begin{equation}
\bar{s}_{d}^{(i)} = \frac{1}{R}\sum_{r=1}^{R} s_{d}^{(i,r)}.
\end{equation}
Then report per-dimension dataset mean:
\begin{equation}
S_d = \frac{1}{N}\sum_{i=1}^{N}\bar{s}_{d}^{(i)}.
\end{equation}
The aggregated benchmark score is
\begin{equation}
S_{\text{overall}} = \frac{1}{3}(S_{\text{IF}} + S_{\text{RQ}} + S_{\text{EE}}).
\end{equation}

To ensure unbiased assessment, videos are assigned to raters using a balanced, randomized sampling strategy. Each video-model pair is evaluated by at least 2 independent raters, and the presentation order of the four methods is shuffled for every trial to eliminate model-specific or sequential bias.

\subsection{Reporting Protocol}

\paragraph{Primary Metrics.}
We report the mean scores for Instruction Following (IF), Rendering Quality (RQ), and Edit Exclusivity (EE) across the entire benchmark. An Overall Score is computed as the unweighted arithmetic mean of these three dimensions, providing a holistic measure of instance removal performance.

\paragraph{Split-wise Analysis.}
To isolate model robustness under varying physical complexities, we additionally report performance partitioned by the Simple and Hard subsets. This granular reporting highlights the "performance decay" models experience when transitioning from basic scenarios to those with strong physical coupling (e.g., reflections and wakes).

\paragraph{Uncertainty and Significance.}
To ensure the reliability of our rankings, we report 95\% Confidence Intervals (CIs) for all primary metrics. For pairwise model comparisons, we employ Bootstrap Significance Testing (with $N=10,000$ iterations). Our analysis confirms that the performance superiority of PISCO-Removal and UniVideo is statistically significant ($p < 0.05$) across all dimensions, particularly on the Hard subset where simpler baselines suffer from severe physical artifacts.

%% file: sec/5_exp.tex
\section{Experiments}
\label{sec:experiments}

\subsection{Experimental Setup}
\paragraph{Benchmark setting.}
We evaluate all methods on the full 95-video benchmark and report:
(1) overall scores,
(2) split-wise scores on Simple and Hard subsets,
(3) qualitative failure analysis.
Unless otherwise noted, all scores are human ratings following \cref{subsec:human_eval}.

\paragraph{Models.}
We benchmark CoCoCo~\cite{zi2025cococo}, UniVideo~\cite{wei2025univideo}, DiffuEraser~\cite{li2025diffueraser}, and PISCO-Removal~\cite{miao2025rose,gao2026pisco}.

\paragraph{Implementation Details.}
All experiments are conducted on a workstation with three \textbf{NVIDIA A100 (80GB)} GPUs. To ensure a high-fidelity evaluation, we process all sequences at a native \textbf{720p (81 frames)} resolution.
The inference latency varies significantly across baselines: \textbf{UniVideo} requires $\sim$3.5 hours per video, while \textbf{PISCO-Removal}, \textbf{DiffuEraser}, and \textbf{CoCoCo} complete in $\sim$30 minutes. 
For models with limited temporal receptive fields (e.g., CoCoCo), we apply a sliding window inference with a \textbf{4-frame overlap} to maintain coherence. 
All outputs are exported in a lossless format to avoid secondary compression artifacts during human evaluation.

\subsection{Main Results}
\label{subsec:main_results}

\begin{table*}[t]
    \centering
    \small
    \caption{Performance comparison on the comprehensive benchmark (95 videos), including overall results and the Simple/Hard split breakdown. Metric scores range from 1 to 4, where higher values indicate superior performance. \textbf{\color{best}Dark green} and \color{second}light green \color{black} denote the best and second-best results within each subset, respectively.}
    \label{tab:comprehensive_results}
    \begin{tabular}{llcccc}
        \toprule
        \textbf{Subset} & \textbf{Method} & \textbf{Instruction Following} $\uparrow$ & \textbf{Rendering Quality} $\uparrow$ & \textbf{Edit Exclusivity} $\uparrow$ & \textbf{Overall} $\uparrow$ \\
        \midrule
        \multirow{4}{*}{\rotatebox{90}{\textbf{Overall}}} 
        & CoCoCo~ & 1.60 & 1.84 & 3.07 & 2.17 \\
        & UniVideo~ & \cellcolor{second}3.06 & \cellcolor{best!20}\textbf{3.45} & \cellcolor{second}3.53 & \cellcolor{second}3.35 \\
        & DiffuEraser~ & 2.89 & 2.63 & 3.52 & 3.01 \\
        & PISCO-Removal~ & \cellcolor{best!20}\textbf{3.62} & \cellcolor{second}3.28 & \cellcolor{best!20}\textbf{3.58} & \cellcolor{best!20}\textbf{3.49} \\
        \midrule
        \multirow{4}{*}{\rotatebox{90}{\textbf{Simple}}} 
        & CoCoCo & 1.75 & 1.98 & 3.33 & 2.35 \\
        & UniVideo & \cellcolor{second}3.21 & \cellcolor{best!20}\textbf{3.34} & 3.53 & \cellcolor{second}3.36 \\
        & DiffuEraser & 2.73 & 2.73 & \cellcolor{best!20}\textbf{3.73} & 3.06 \\
        & PISCO-Removal & \cellcolor{best!20}\textbf{3.75} & \cellcolor{second}3.32 & \cellcolor{second}3.57 & \cellcolor{best!20}\textbf{3.55} \\
        \midrule
        \multirow{4}{*}{\rotatebox{90}{\textbf{Hard}}} 
        & CoCoCo & 1.52 & 1.76 & 2.92 & 2.07 \\
        & UniVideo & 2.96 & \cellcolor{best!20}\textbf{3.53} & \cellcolor{second}3.53 & \cellcolor{second}3.34 \\
        & DiffuEraser & \cellcolor{second}3.00 & 2.56 & 3.38 & 2.98 \\
        & PISCO-Removal & \cellcolor{best!20}\textbf{3.45} & \cellcolor{second}3.23 & \cellcolor{best!20}\textbf{3.59} & \cellcolor{best!20}\textbf{3.42} \\
        \bottomrule
    \end{tabular}
    \vspace{3mm}
\end{table*}

\paragraph{Overall comparison.}

\cref{tab:comprehensive_results} summarizes the overall performance. We observe a clear performance hierarchy: \textbf{PISCO-Removal} and \textbf{UniVideo} consistently define the state-of-the-art across all metrics, forming a high-fidelity tier. In contrast, while \textbf{DiffuEraser} provides competitive efficiency, it suffers from a significant "fidelity gap" compared to the leaders. The most striking finding is the universal performance decay on the \emph{Hard} subset, where even the top-performing models struggle to maintain physical consistency, highlighting the diagnostic value of our benchmark.

\paragraph{Instruction following.}
This dimension evaluates the model's ability to ground textual instructions into pixel-level removal, where we observe a clear trade-off between semantic autonomy and execution reliability. \textbf{UniVideo} and \textbf{PISCO-Removal}, as representative end-to-end architectures, demonstrate superior visual integration in the majority of cases; however, they exhibit occasional "grounding drift" where the target subject is either ignored or only partially erased. This suggests that while their latent semantic alignment is powerful, it can occasionally fail to localize the instance accurately amidst cluttered backgrounds. In contrast, \textbf{DiffuEraser} achieves the highest reliability in complete removal across the benchmark. Since its pipeline is explicitly constrained by the input mask, it bypasses the semantic localization errors inherent in end-to-end models, ensuring the designated regions are always processed. Notably, \textbf{CoCoCo} consistently fails this dimension; as a general-purpose inpainting model, it lacks the specialized inductive bias for large-scale instance removal, often producing "ghosting" artifacts that retain the original object's silhouette or failing to initiate the removal command altogether in favor of local texture synthesis.

\paragraph{Rendering quality.}
This dimension evaluates visual fidelity and temporal stability, where we observe a stark contrast in spatial resolution and coherence. \textbf{UniVideo} and \textbf{PISCO-Removal} consistently produce the most visually pleasing results, maintaining high-frequency textures that blend seamlessly with the original background. While they exhibit occasional flickering in complex dynamic scenes, their overall rendering remains stable at 720p. In contrast, \textbf{DiffuEraser} suffers from pervasive \emph{spatial blurring} within the inpainted regions. Despite its ability to reliably remove the target, the synthesized textures often lack the sharpness of the surrounding environment, creating a noticeable "patchwork" effect that disrupts the scene's visual harmony. \textbf{CoCoCo} performs the worst in this category, frequently generating severe "ghosting" artifacts—where remnants of the original object reappear as semi-transparent textures—and significant temporal flickering. These failures suggest that while general inpainting models can fill small holes, they lack the structural priors necessary to reconstruct large-scale, 81-frame backgrounds with the requisite physical and temporal plausibility.

\paragraph{Edit exclusivity.}
This dimension assesses the models' precision in localized editing, specifically focusing on whether the transformation "leaks" into non-target background regions. Overall, all four baselines demonstrate a respectable ability to preserve the surrounding scene context. \textbf{UniVideo} and \textbf{PISCO-Removal} exhibit high spatial fidelity, maintaining the original pixel values of the static background with minimal drift. However, \textbf{DiffuEraser} occasionally suffers from \emph{blurring leakage}, where the spatial smoothing intended for the erased region inadvertently spreads beyond the mask boundaries into the neighboring textures. This creates a subtle but perceptible halo of reduced sharpness in the background, a phenomenon that is particularly visible at 720p. For the \emph{Hard} subset involving complex reflections, we observe that models often struggle to disentangle the target instance from its environmental "side effects," sometimes leading to unintended modifications of the global lighting or shadow maps in areas that should remain untouched.

\subsection{Simple vs. Hard Split Analysis}
\label{subsec:split_analysis}

The split-wise analysis isolates model robustness against physics-coupled interactions. As detailed in \cref{tab:comprehensive_results}, we observe a noticeable performance degradation across multiple dimensions when transitioning from the \emph{Simple} to the \emph{Hard} subset. Looking beyond the aggregate scores, the multi-dimensional breakdown reveals that the most significant drops typically occur in \textbf{Instruction Following} and \textbf{Rendering Quality}. For instance, top-tier models like \textbf{PISCO-Removal} experience a drop in IF from 3.75 on the Simple set to 3.45 on the Hard set, illustrating the increased difficulty of executing clean removals amidst complex physical constraints. While all four baselines can reliably erase an isolated object against a static background in the \emph{Simple} set, they frequently struggle to maintain structural consistency when the task scales in physical complexity.

The interaction types involving \textbf{specular reflections} and \textbf{dynamic wakes/fluid ripples} prove to be the most challenging. In many \emph{Hard} cases, even when the primary instance is successfully erased by leaders like \textbf{UniVideo} or \textbf{PISCO-Removal}, its corresponding physical "side effects"—such as a moving shadow on a textured wall or a mirror reflection on a car's surface—remain stubbornly visible. This severely impacts the \textbf{Edit Exclusivity} scores for methods like \textbf{DiffuEraser}, which drops from 3.73 (Simple) to 3.38 (Hard). Its local blurring strategy fails to properly propagate the background's global illumination, leading to physically implausible "ghost reflections." These findings underscore that current Video Instance Removal (VIR) models generally treat removal as a 2D texture-filling task rather than a 3D-aware scene reconstruction, highlighting a critical direction for future physics-augmented research.
\begin{figure*}[t]
    \centering
    \includegraphics[width=\linewidth]{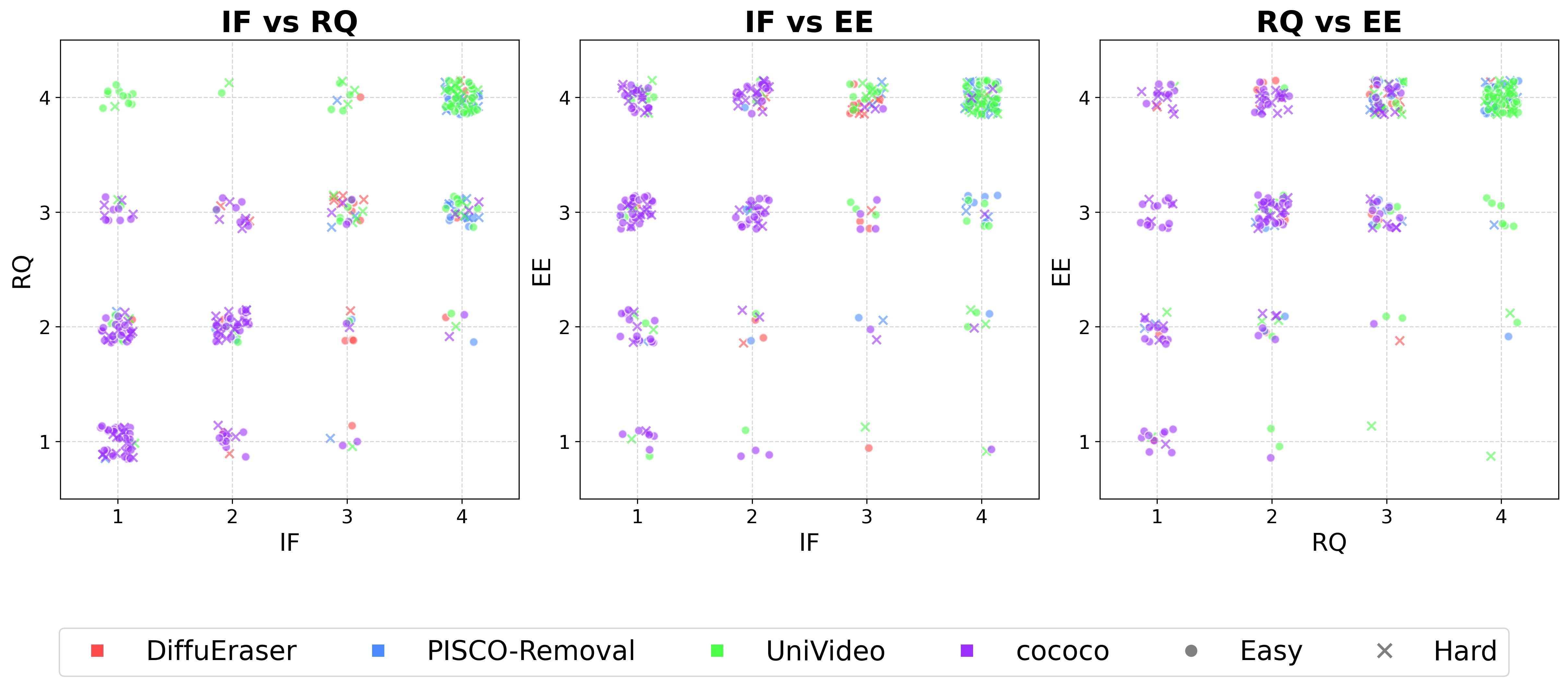}
    \caption{\textbf{Cross-metric trade-off analysis across Instruction Following (IF), Rendering Quality (RQ), and Edit Exclusivity (EE).} Each point represents a single video-model pair. Markers distinguish between the \emph{Simple} ($\bullet$) and \emph{Hard} ($\times$) subsets. The tight clustering in the top-right quadrant for \textbf{PISCO-Removal} and \textbf{UniVideo} indicates high-fidelity, balanced performance. Conversely, the wide dispersion of \textbf{DiffuEraser} and \textbf{CoCoCo} reveals inherent trade-offs between reliable target removal and background preservation.}
    \label{fig:tradeoff}
\end{figure*}

\subsection{Cross-Metric Trade-off Analysis}

To analyze whether methods sacrifice one dimension for another, we examine the pairwise relationships in Fig.~\ref{fig:tradeoff}. This visualization exposes the inherent tension between a model's ``semantic fluidity'' and its adherence to local constraints.

\paragraph{High-Fidelity Clustering.} \textbf{PISCO-Removal} and \textbf{UniVideo} (blue and green) exhibit tight clustering in the (4,4) quadrants of the IF-RQ and IF-EE plots, defining a high-fidelity tier that balances task completion with background preservation. However, a clear shift toward lower RQ scores is evident for the \textbf{Hard} subset ($\times$), confirming that complex physical coupling remains the primary performance bottleneck even for top-tier models.

\paragraph{Reliability vs. Precision.} \textbf{DiffuEraser} (red) achieves high IF scores (Score 3--4) due to its mask-guided nature, yet shows significant dispersion on the RQ and EE axes. This reflects a trade-off where reliable object removal often introduces spatial blurring or ``halo'' artifacts in non-target regions. Conversely, \textbf{CoCoCo} (purple) is heavily skewed toward low IF scores (Score 1--2), often failing to execute the removal command while maintaining high EE scores simply by leaving the scene unedited.

\paragraph{Metric Correlations.} Statistical analysis reveals that IF and RQ are moderately correlated ($r \approx 0.65$), suggesting that models with better instruction grounding typically synthesize more plausible textures. However, the weak correlation between RQ and EE in lower-performing models underscores the necessity of our decoupled protocol: visual plausibility alone does not guarantee that the rest of the scene remains untouched.

% \begin{figure*}[t]
%     \centering
%     \fbox{\rule{0pt}{2.25in}\rule{0.98\linewidth}{0pt}}
%     \caption{Failure atlas grouped by physics interaction type (reflection, specular coupling, long-range occlusion, and background drift).}
%     \label{fig:failure_atlas}
% \end{figure*}

% \subsection{Practical Recommendation for Benchmark Users}

% To improve comparability across future submissions, we recommend reporting all three dimensions and split-wise metrics by default.In addition, each submission should provide at least one qualitative panel with explicit failure labels to avoid over-reliance on single-number ranking and help identify which dimension is improved by a method update. For formal benchmark inclusion, we propose a standardized checklist: (1) Consistency in Resolution, requiring all sequences to be rendered at 720p to prevent artifacts from being masked by downsampling; (2) Hardware Disclosure, reporting GPU hours (e.g., A100 usage) to contextualize the efficiency-quality trade-off; and (3) Mask Transparency, specifying whether masks are ground-truth or model-generated. The PVIR leaderboard will prioritize performance on the \emph{Hard} subset as the primary ranking metric. To maintain scientific integrity, all entries must provide reproducible inference scripts or Dockerized environments, and authors are encouraged to document systemic failures in optical or fluid interactions as part of their submission to foster community-wide understanding of current physical limitations.

\subsection{Discussion}

\paragraph{Why decoupled evaluation matters.} 
Traditional VIR evaluation often relies on a single visual plausibility score, which masks critical failure modes. Our three-axis protocol reveals that a model can score highly on \emph{instruction following} while simultaneously failing \emph{edit exclusivity}, or produce visually plausible textures while completely missing the target removal. By decoupling these dimensions, we expose the inherent trade-offs in current architectures—such as the balance between the strict spatial constraints of mask-guided models like \textbf{DiffuEraser} and the semantic fluidity of end-to-end generative models like \textbf{UniVideo}. This granular mapping is essential for diagnosing whether a model's failure stems from poor instruction grounding, low rendering fidelity, or unintended scene drift.

\paragraph{Current limitations.} 
Current benchmark evaluations remain heavily human-centered, making them both cost-sensitive and difficult to scale. While we provide a comprehensive human study, the development of robust automatic metrics for physics-aware side effects—such as detecting residual reflections, lingering shadows, or fluid inconsistencies—remains highly challenging. Existing feature-space proxies fail to capture this high-level physical logic, indicating an urgent need for automated, physics-aware evaluation models.

\paragraph{Future directions.} 
Our empirical results highlight a clear paradigm shift: large-scale generative models significantly outperform smaller, traditional inpainting networks. Furthermore, within the regime of large models, specialized architectures fine-tuned for precise editing demonstrate noticeable superiority over unified, general-purpose video models in handling strict spatial constraints. This suggests that while foundational priors are necessary, they are not sufficient for instance-level physical accuracy. Consequently, future advancements in video instance removal should focus on two key pillars: the continued scaling of large video foundation models, and the rigorous curation of high-quality, domain-specific datasets designed to inject precise physical and spatial constraints into these models.

\paragraph{Benchmark extension roadmap.} 
As an evolving platform, the PVIR roadmap includes expanding the dataset to encompass greater category diversity, extended sequences (e.g., 10+ seconds) to stress-test long-term temporal consistency, and more complex physical phenomena such as multi-object interactions and fluid/smoke dynamics.

%% file: sec/6_con.tex
\section{Conclusion}
\label{sec:conclusion}

In this work, we introduced the Physics-Aware Video Instance Removal (PVIR) benchmark, a dedicated evaluation infrastructure designed to bridge the gap between visual plausibility and physical consistency in video editing. Our benchmark contributes three key pillars: a high-quality 95-video dataset with dense annotations, a Simple/Hard difficulty split driven by complex physical interactions, and a decoupled evaluation protocol spanning instruction following, rendering quality, and edit exclusivity. By benchmarking four representative models under a unified, high-resolution inference setup, we revealed a significant performance hierarchy and a universal ``physics blindness'' in current state-of-the-art methods.

Crucially, our analysis demonstrates that while existing models excel in static background synthesis, they suffer from severe performance when encountering optical reflections, shadows, and fluid interactions in our Hard subset. This finding underscores that current video instance removal is still predominantly treated as a 2D texture-filling task rather than a 3D-aware physical reconstruction. We offer PVIR as a rigorous yardstick to encourage the community to move beyond surface-level aesthetic metrics and toward physically grounded video intelligence.

%% file: sec/X_suppl.tex
\clearpage
\appendix
\onecolumn

\section{Supplementary Material}

\subsection{Evaluation Guide for Human Raters}
\label{app:rating_guide}

This section provides the detailed 1--4 criteria used by annotators.
All dimensions are strictly independent.

\paragraph{Instruction Following (IF).}
\textbf{Core question:} Does the edited video correctly satisfy the removal instruction?
\begin{itemize}[leftmargin=1.2em]
    \item \textbf{4}: Perfect adherence. Target instance is removed cleanly with no obvious remnants.
    \item \textbf{3}: High adherence. Target removed with minor residual imperfections.
    \item \textbf{2}: Low adherence. Target partly removed, or obvious residual traces remain.
    \item \textbf{1}: Failure. Target remains or output is largely unrelated to the instruction.
\end{itemize}

\paragraph{Rendering Quality (RQ).}
\textbf{Core question:} Is the filled content visually plausible and temporally stable?
\begin{itemize}[leftmargin=1.2em]
    \item \textbf{4}: Excellent quality, negligible artifacts, temporally stable.
    \item \textbf{3}: Acceptable quality, moderate artifacts but viewable.
    \item \textbf{2}: Poor quality, obvious distortion/flicker affecting readability.
    \item \textbf{1}: Unusable quality, severe artifacts or physical implausibility.
\end{itemize}

\paragraph{Edit Exclusivity (EE).}
\textbf{Core question:} Are non-target regions preserved?
\begin{itemize}[leftmargin=1.2em]
    \item \textbf{4}: Non-target regions are effectively unchanged.
    \item \textbf{3}: Minor non-target drift but scene remains mostly preserved.
    \item \textbf{2}: Noticeable over-editing in multiple non-target areas.
    \item \textbf{1}: Major uncontrolled edits; scene is substantially altered.
\end{itemize}